\documentclass[10pt,twocolumn,letterpaper]{article}

\usepackage[pagenumbers]{cvpr} 

\usepackage{float} 

\definecolor{cvprblue}{rgb}{0.21,0.49,0.74}
\usepackage[pagebackref,breaklinks,colorlinks,allcolors=cvprblue]{hyperref}

\usepackage[table]{xcolor}
\definecolor{lightgray}{gray}{0.9}
\usepackage[accsupp]{axessibility}  

\def\paperID{} 
\def\confName{CVPR}
\def\confYear{2026}

\title{Enhancing Part-Level Point Grounding for Any Open-Source MLLMs}

\newcommand{\Hquad}{\hspace{0.5em}} 
\author{
    Jin-Cheng Jhang$^{1}$ \Hquad
    Fu-En Wang$^{2}$ \Hquad
    Xin Yang$^{2}$ \Hquad
    Nan Qiao$^{2}$ \Hquad
    Lu Xia$^{2}$ \Hquad
    \\
    Min Sun$^{1,2}$ \Hquad
    Cheng-Hao Kuo$^{2}$ \Hquad
    \\
    $^{1}$National Tsing Hua University \Hquad 
    $^{2}$Amazon
    \\
    {\tt\small frank890725@gapp.nthu.edu.tw} \Hquad
    {\tt\small sunmin@ee.nthu.edu.tw}
    \\
    {\tt\small \{fuenwang, yngxin, qiaonan, luxial, chkuo\}@amazon.com
    }
}

\begin{document}
\maketitle
\begin{abstract}
Visual grounding aims to associate free-form textual queries with specific regions in an image. While recent Multimodal Large Language Models (MLLMs) have demonstrated promising capabilities in this domain, they primarily excel at object-level grounding and often struggle with part-level grounding—an essential requirement for fine-grained tasks such as robotic manipulation. In this work, we introduce a general approach that equips any open-source MLLMs with accurate 2D part-level point grounding, offering a more direct alternative to conventional grounding representations. Our method leverages the attention mechanisms inherently present in MLLMs. By synthesizing text-conditioned, grounding-aware queries within intermediate layers via the proposed Q-Synth Module, we capture target-relevant attention patterns and refine them with a lightweight Attention-to-Point Decoder, which converts these patterns into a point-centric heatmap for final prediction. Notably, all original MLLM parameters are frozen, ensuring full preservation of their pre-trained capabilities. Experiments show that our design consistently improves part-level grounding accuracy across datasets and can be seamlessly integrated into any open-source MLLMs.
\end{abstract}    
\section{Introduction}
\label{sec:intro}

Point grounding aims to associate a textual description with a specific point in an image, providing a direct cue that links semantic concepts to precise spatial locations. Beyond vision-language understanding, this point-based formulation may also serve as a useful perception primitive for downstream robotic interaction~\cite{manuelli2019kpam, qin2020keto, huang2024rekep, motoda2025suctionprompt, fangandliu2024moka, yuan2024robopoint}.

\begin{figure}[tp]
    \centering
    \includegraphics[width=\linewidth]{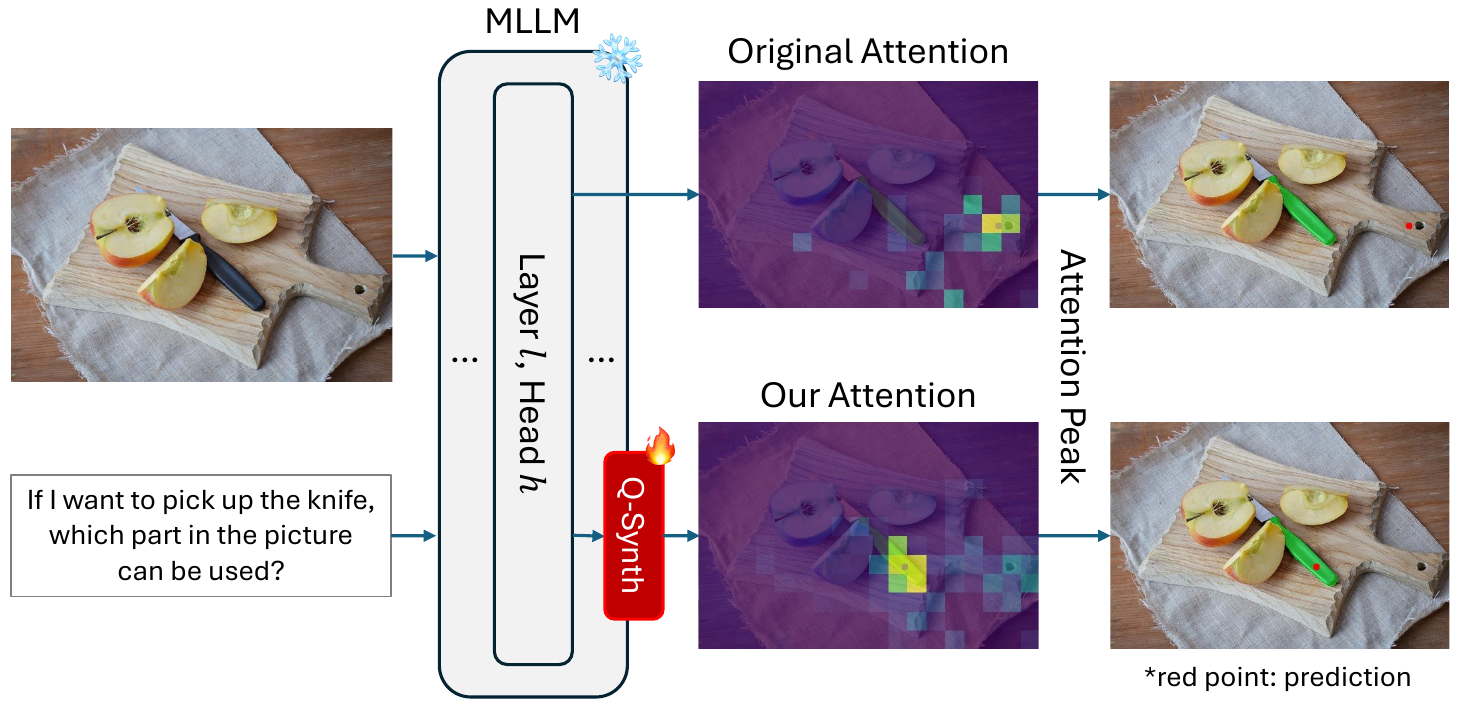}
    \caption{Overview of the core idea.
    We aim to synthesize grounding-aware queries from a frozen MLLM via the proposed Query Synthesis (Q-Synth) Module to extract more accurate attention patterns for part-level point grounding. 
    Warmer colors (yellow) in the attention maps indicate higher attention values. The red points denote the predicted grounding locations, and the green masks show the corresponding ground-truth regions.}
    \label{fig:teaser}
\end{figure}

For instance, in a garment-folding scenario, localizing a point on parts such as the cuff, collar, or hem at each step may provide informative cues for where the garment could be grasped or placed.
More broadly, this example highlights the limitation of object-level grounding for physical interaction tasks, where fine-grained manipulation often depends on localizing a specific object part rather than the object as a whole.
This makes \textit{part-level} grounding—such as identifying the handle of the knife shown in Figure~\ref{fig:teaser}—particularly relevant.
Moreover, compared with conventional grounding formats such as bounding boxes or segmentation masks, point-based grounding provides a more direct and compact representation that is naturally aligned with action-relevant locations.

In recent years, Multimodal Large Language Models (MLLMs) have demonstrated remarkable reasoning abilities, enabling them to decompose long-horizon robotic tasks into manageable subtasks~\cite{driess2023palm, wu2024mldt, yang2025guiding, li2025towards}.
If such capable MLLMs could also acquire accurate visual grounding abilities, they would not only understand \textit{what} steps to take but also \textit{where} to execute them based on their visual observations.
Indeed, some recent MLLMs~\cite{deitke2025molmo, bai2025qwen25vltechnicalreport, team2025gemini} have recognized the importance of point-level grounding and explicitly included it in their learning objectives. 
However, as shown in Table~\ref{tab:obj_level}, existing models perform well on object-level grounding but still struggle with part-level targets.

%
\begin{table}
  \caption{Comparison of object-level and part-level point grounding on the PACO~\cite{ramanathan2023paco} dataset.
  The models generate point coordinates in text form, referred to as text pointing in later experiments.
  Acc. denotes accuracy, measured by whether the predicted point lies within the ground-truth mask.
  First-Gen-MLLM refers to a model ($<$10B parameters) not trained for point grounding.
  The results show that while existing models perform well on object-level grounding, they still struggle with precise part-level localization.}
  \label{tab:obj_level}
  \centering
    \begin{tabular}{lcc}
    \toprule
     & Obj-level Acc. & Part-level Acc. \\
    \midrule        
    Molmo-7B~\cite{deitke2025molmo}   & 0.854 & 0.487\\
    Qwen2.5-VL-7B~\cite{bai2025qwen25vltechnicalreport}   & 0.838 & 0.407\\
    \midrule
    First-Gen-MLLM   & 0.155 & 0.068 \\
    \bottomrule
    \end{tabular}
\end{table}
Therefore, how to enhance the existing grounding ability of MLLMs has become a popular and crucial research direction.
Several prior works~\cite{lai2024lisa, pi2024perceptiongpt, rasheed2024glamm, ren2024pixellm, wei2024lasagna} have attempted to equip MLLMs with segmentation capabilities by fine-tuning them to output special tokens (e.g., \texttt{[SEG]}) and then decoding the segmentation masks from these tokens. 
However, such fine-tuning often causes the models to overfit to the special token outputs, degrading their original reasoning and natural conversational abilities, as observed in~\cite{wu2025flmm}.
To overcome this limitation, recent studies~\cite{kang2025your, wu2025flmm} have discovered an emerging property within the attention layers of MLLMs: certain text-to-image attention maps inherently capture strong and spatially consistent signals that reveal what the model is focusing on. These attention patterns can thus be leveraged to infer grounding results without modifying the original language modeling head.

Although directly using the native attention maps offers a feasible way to obtain grounding results from MLLMs, it remains suboptimal in accuracy—particularly for part-level grounding, which demands highly precise spatial localization. 
Building on prior findings, we propose a learnable approach that refines native attention patterns for more accurate grounding, as illustrated in Figure~\ref{fig:teaser}.
Specifically, we introduce a \textbf{Query Synthesis (Q-Synth) Module} that takes a sequence of text-query features and iteratively aggregates salient semantics to synthesize a single grounding-aware query, which drives a targeted text-to-image attention map.
To further achieve higher spatial resolution and finer localization, we design an \textbf{Attention-to-Point (A2P) Decoder} that performs progressive upsampling and spatial refinement to generate the final heatmap for precise point grounding.
In addition, we introduce an \textbf{SDF-based penalty field} that provides point-centric supervision and guides the model to form sharper, more spatially coherent heatmaps.
%
%
Experiments demonstrate that our method can be seamlessly integrated into various open-source MLLMs and consistently enhances their point grounding performance across different benchmarks. 
Remarkably, these improvements are achieved without compromising the models’ original reasoning and generalization abilities.

To sum up, our main contributions are as follows:
\begin{itemize}
    \item We identify part-level point grounding as an important perception primitive and present a general framework that equips any open-source MLLMs with this ability while preserving their pre-trained capabilities.
    \item We design a lightweight attention refinement module that integrates query synthesis, attention decoding, and an SDF-based penalty field to produce grounding-aware attention patterns for precise point localization.
    \item Experiments across multiple MLLMs show that our method integrates seamlessly with existing architectures and consistently improves grounding performance without additional fine-tuning.
\end{itemize}
\section{Related Works}
\label{sec:related_works}

\subsection{Point-based Grounding}
Point grounding represents visual targets as spatial points conditioned on language instructions.
Recent works have demonstrated the effectiveness of point grounding in real-world scenarios.
Rekep~\cite{huang2024rekep} and MOKA~\cite{fangandliu2024moka} generate multiple actionable keypoint candidates and query an MLLM (e.g., GPT-4o~\cite{achiam2023gpt}) via visual prompting to identify relationships or attributes among them, thereby guiding the robot end-effector trajectories for various manipulation tasks.
Similarly, SuctionPrompt~\cite{motoda2025suctionprompt} provides suction-point candidates to an MLLM, which identifies the optimal grasping location and achieves reliable performance in convenience store scenarios.
Other than \textit{selecting points} from candidates, RoboPoint~\cite{yuan2024robopoint} fine-tunes MLLMs directly to become pointing specialists, demonstrating the versatility of point grounding across tasks such as manipulation, augmented reality, and navigation.

Meanwhile, several recent MLLMs~\cite{deitke2025molmo, bai2025qwen25vltechnicalreport, team2025gemini} have begun incorporating point grounding into their training objectives, allowing native text-based pointing outputs. 
However, these models typically perform well at object-level grounding but struggle with part-level localization, which demands finer spatial precision. 
Moreover, models lacking explicit point-grounding supervision—referred to as First-Gen-MLLMs—remain unable to generate meaningful pointing results.
Therefore, our work aims to enhance the part-level point grounding ability of any MLLMs, enabling more precise and actionable grounding for a broader range of robotic applications.

\begin{figure*}[tp]
    \centering
    \includegraphics[width=\linewidth]{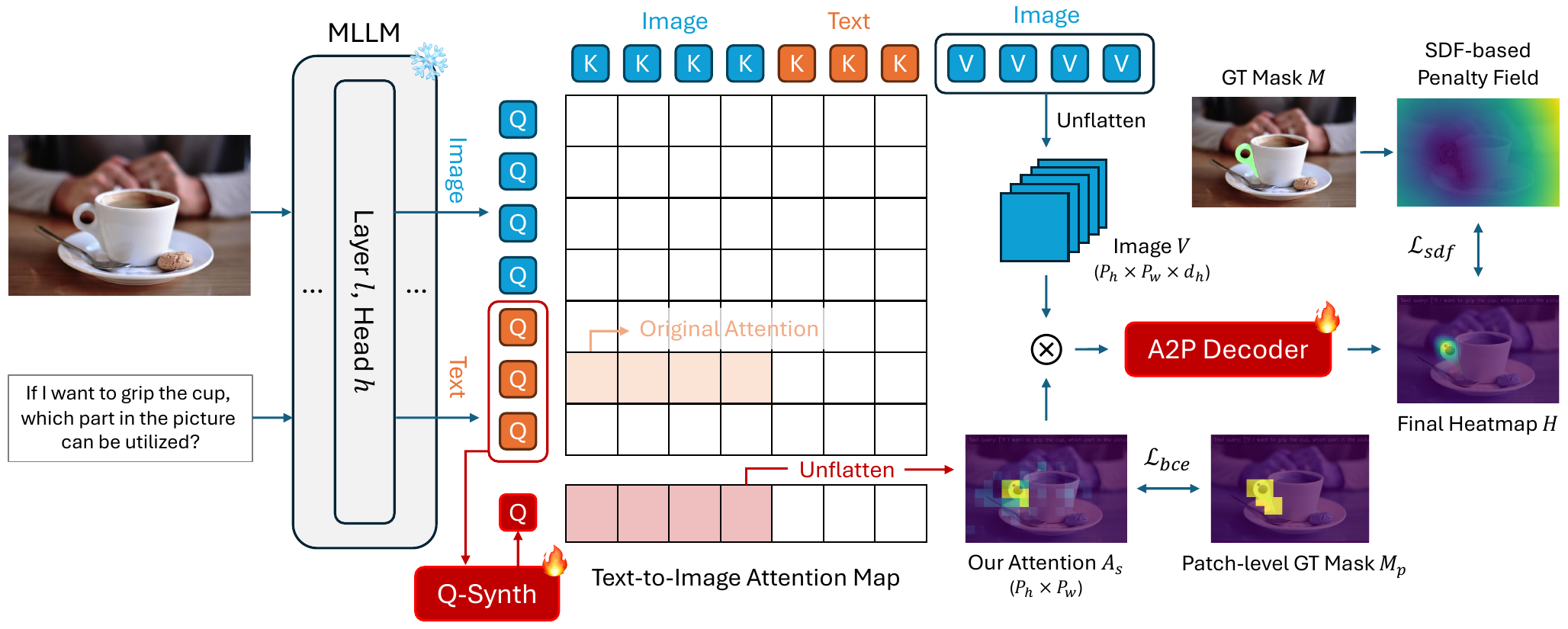}
    \caption{Overview of our framework. 
    Instead of relying on the native text-to-image attention, the proposed Query Synthesis (Q-Synth) Module (Sec.~\ref{sec:qs_module}) conditions on all text features and synthesizes a single grounding-aware query $q_s$ to extract a refined attention map $A_s$. 
    The Attention-to-Point (A2P) Decoder (Sec.~\ref{sec:a2p}) then combines $A_s$ with image $V$ to form attention-weighted features and decodes a high-resolution heatmap $H$ for point prediction. 
    For training, we use a per-patch BCE loss to guide $A_s$ and an SDF-based penalty field to shape $H$ (Sec.~\ref{sec:loss}). 
    In the penalty field, lower values (shown in cooler colors) indicate the target regions.}
    \label{fig:framework}
\end{figure*}

\subsection{Enhance Grounding Ability for MLLMs} 
Recent studies on enhancing the grounding ability of Multimodal Large Language Models (MLLMs) generally follow two main directions.
The first direction focuses on fine-tuning the parameters of MLLMs to explicitly equip them with grounding capabilities. 
Several works~\cite{chen2023shikra, peng2023kosmos, you2023ferret} train MLLMs to autoregressively output bounding-box coordinates in text form, while others~\cite{lai2024lisa, pi2024perceptiongpt, rasheed2024glamm, ren2024pixellm, wei2024lasagna} introduce special grounding tokens (e.g., \texttt{[SEG]}) and decode boxes or masks from them. Although effective, these methods require extensive training and often degrade the model’s native reasoning and conversational abilities.

The second direction explores frozen-parameter approaches, leveraging the inherent attention mechanisms of MLLMs without modifying their pre-trained weights. Recent works~\cite{wu2025flmm, kang2025your} reveal an emergent grounding property within the attention patterns of MLLMs, where text-to-image attention maps can naturally highlight the visual regions relevant to a given text query. By extracting and interpreting these attention maps, grounding results can be obtained while fully preserving the model’s original multimodal reasoning capabilities.
Building upon this observation, we learn to refine and extract more accurate attention patterns, further enhancing grounding performance.
\section{Preliminary}
\label{sec:preliminary}


\paragraph{MLLM Structure.}
A Multimodal Large Language Model (MLLM) for image--text input typically consists of a vision encoder, a projector, and a Large Language Model (LLM).
Given an image $I \in \mathbb{R}^{H \times W \times 3}$, the vision encoder divides $I$ into a grid of patches and produces visual embeddings $V \in \mathbb{R}^{P_h \times P_w \times d_v}$, where $P_h$ and $P_w$ denote the height and width of the patch grid, and $d_v$ is the visual feature dimension.
A projector maps $V$ into the LLM feature space, yielding a sequence of visual tokens $X_v \in \mathbb{R}^{P_h P_w \times d}$, where $d$ is the LLM hidden size.
Let $X_t \in \mathbb{R}^{L \times d}$ denote the tokenized text embeddings, where $L$ is the number of text tokens.
The multimodal sequence is formed by concatenation, $X = [X_v; X_t] \in \mathbb{R}^{(P_h P_w + L) \times d}$, and is fed into the LLM for autoregressive generation.
\paragraph{Text-to-Image Attention Maps.}
In MLLMs, text and image features typically interact through attention mechanisms within the LLM. 
In layer $l$ and attention head $h$, text features act as queries $Q_t$, while image features serve as keys $K_v$ and values $V_v$. 
For a specific target text query vector $q_{\mathrm{tg}}$, its attention weights $a$ are computed as
\begin{equation}
    a^{l,h} = \mathrm{softmax}\!\left(\frac{q_{\mathrm{tg}} K_v^{\top}}{\sqrt{d_h}}\right) \in \mathbb{R}^{P_h P_w + L},
    \label{attn_w}
\end{equation}
where $d_h$ is the dimension of each attention head, typically $d$ divided by the number of heads. 
The text-to-image attention component can be obtained by taking the first $P_h P_w$ elements, i.e., $a^{l,h}[:P_h P_w]$, which can be reshaped into a 2D attention map $A^{l,h} \in \mathbb{R}^{P_h \times P_w}$.

\paragraph{Localization Heads.}
Recent work~\cite{kang2025your} shows that only a small subset of attention heads within LLM layers is crucial for visual grounding and proposes a method to select these heads. 
By aggregating the text-to-image attention maps from the selected heads as spatial cues, MLLMs can produce bounding-box or segmentation-mask outputs in a zero-shot manner. 
This phenomenon is observed across different MLLMs. 
Following this finding, we first identify the top $k$ grounding-relevant heads and use them as the basis for our method.
\section{Method}
\label{sec:method}
Our proposed framework is illustrated in Figure~\ref{fig:framework}. 
It consists of two main components: the Query Synthesis Module (Sec.~\ref{sec:qs_module}) and the Attention-to-Point Decoder (Sec.~\ref{sec:a2p}). 
In addition, we introduce tailored training objectives to jointly optimize both modules in an end-to-end manner (Sec.~\ref{sec:loss}).

\subsection{Query Synthesis (Q-Synth) Module} \label{sec:qs_module}

In~\cite{kang2025your}, the target text token $q_{\mathrm{tg}}$ is always chosen as the last token to represent the entire sentence, following the autoregressive property of the LLM. 
For example, in the sentence ``\textit{Point to the handle of the knife.}'', the query vector of the last token [.] is used in their experiments. 
However, the last token may not always capture the complete semantics of the sentence, especially when it involves multiple concepts or requires further reasoning.

\begin{figure}[tp]
    \centering
    \includegraphics[width=\linewidth]{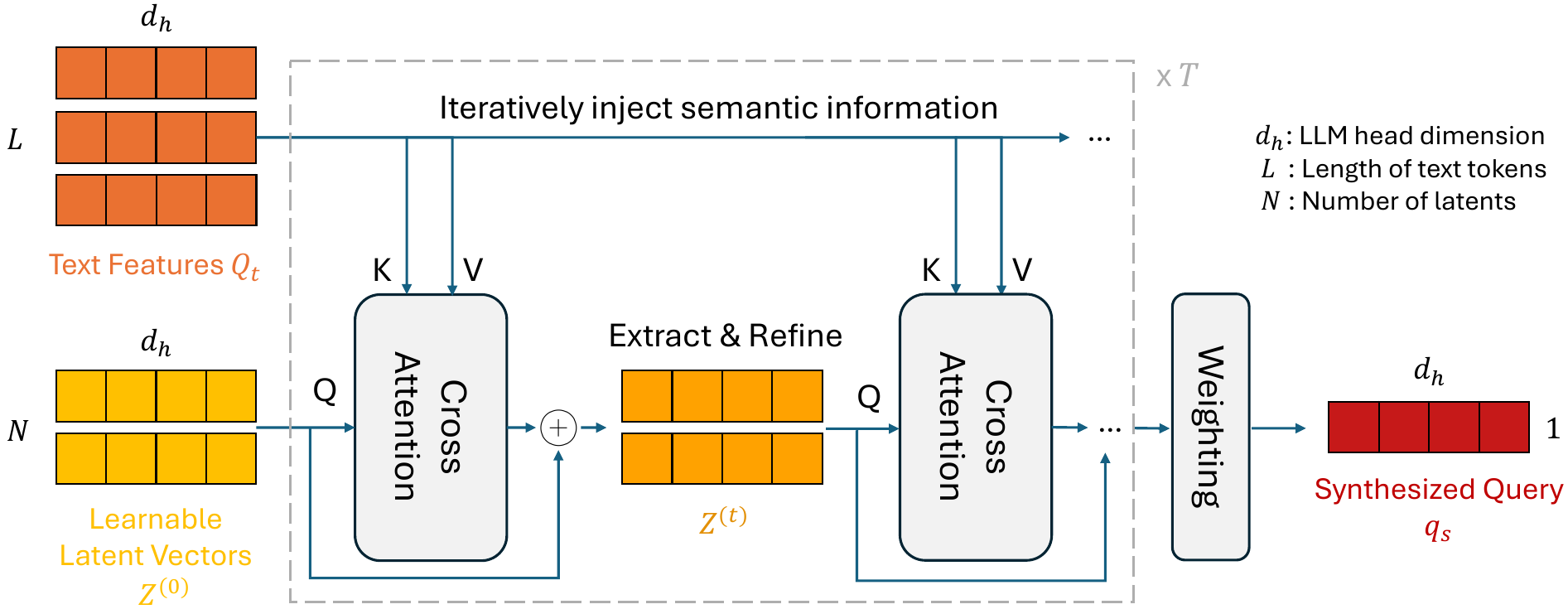}
    \caption{Overview of the Query Synthesis (Q-Synth) Module. 
    Text features $Q_t$ from the LLM and $N$ learnable latent vectors $Z^{(0)}$ are processed by stacked cross-attention layers, where $Q_t$ serves as keys/values to iteratively inject semantics while the latents $Z^{(t)}$ extract and refine the most relevant text information.
    After $T$ iterations, a lightweight weighting network aggregates the refined latents to produce a single query $q_s$ for grounding-aware attention extraction.}
    \label{fig:q-synth}
\end{figure}

To address this limitation, we propose the Query Synthesis (Q-Synth) Module, as illustrated in Figure~\ref{fig:q-synth}, which conditions on all text queries $Q_t \in \mathbb{R}^{L \times d_h}$ from LLM layer $l$ and head $h$ to synthesize a single query $q_s \in \mathbb{R}^{1 \times d_h}$ for more comprehensive and grounding-aware attention map extraction. 
The module employs a stack of cross-attention layers that iteratively model the interaction between the text features and a set of learnable latent queries. 
Specifically, we initialize $N$ learnable latent vectors $Z^{(0)} \in \mathbb{R}^{N \times d_h}$ as the queries, and use the text features as both keys and values, i.e., $K_s = V_s = Q_t$. 
Through multiple rounds of cross-attention ($t = 1, \dots, T$), the text features iteratively inject rich semantic information, while the latent queries extract and summarize the most relevant semantics:
\begin{equation}
    Z^{(t)} = Z^{(t-1)} + \mathrm{CrossAttn}\!\left(Z^{(t-1)},\, K_s,\, V_s\right),
    \label{eq:qs_cross}
\end{equation}
where $Z^{(t)} \in \mathbb{R}^{N \times d_h}$ denotes the refined latent set at layer $t$. 
After $T$ refinement layers, we obtain $Z^{(T)} = [z_1, \dots, z_N]$. 
A lightweight multilayer perceptron (MLP) then computes importance weights over these embeddings to synthesize the final query:
\begin{equation}
    q_s = \sum_{i=1}^{N} \alpha_i z_i, 
    \quad 
    \alpha_i = \mathrm{softmax}\!\left(\mathrm{MLP}(z_i)\right),
    \label{qs_eqn}
\end{equation}
where $\alpha_i$ represents the learned weighting of each latent. 
The resulting $q_s$ is used to compute our synthesized text-to-image attention map $A_s$, similar to Eq.~\ref{attn_w}, serving as a replacement for $q_{\mathrm{tg}}$.

To make $q_s$ grounding-aware and aligned with the visual features, the Q-Synth Module is trained using a binary cross-entropy (BCE) loss, as detailed in Sec.~\ref{sec:loss}.
Besiedes, since we select the top $k$ attention heads as localization heads, a separate Q-Synth Module is attached to each of them, producing $k$ synthesized attention maps. 
These maps are designed to highlight the target regions corresponding to the input text description and are subsequently ensembled in Sec.~\ref{sec:a2p}.


\subsection{Attention-to-Point (A2P) Decoder}
\label{sec:a2p}

Due to the patchification of images in the vision encoder, the resolution of the extracted attention maps is relatively low (typically, $P_h$ and $P_w$ are the original image dimensions divided by 14). 
Even if the correct patch is identified from the attention map, directly pointing to the patch center often fails to yield accurate localization.

To overcome this, we introduce the Attention-to-Point (A2P) Decoder, which produces a higher-resolution, point-focused heatmap for final point prediction. 
Given the synthesized attention map $A_s$ from the Q-Synth Module, derived from $q_s$ and the image features $K_v$, we further leverage the original image $V_v$ for decoding, as it contains the actual visual content represented in the LLM attention head. 
We obtain a spatially modulated feature map $F \in \mathbb{R}^{P_h \times P_w \times d_h}$ by weighting the image $V_v$ with the synthesized attention map $A_s$, effectively highlighting the target regions while suppressing irrelevant areas.

Since we have $k$ synthesized attention maps from different localization heads, a lightweight MLP learns to weight these $k$ feature maps. 
The weighted maps are concatenated along the feature dimension and passed through a $1\times1$ convolution to reduce dimensionality, producing fused feature maps $F_{\mathrm{fused}} \in \mathbb{R}^{P_h \times P_w \times d_{\mathrm{fused}}}$. 
Finally, a sequence of convolutional layers interleaved with bilinear upsampling operations performs spatial refinement and progressive upscaling, generating the final high-resolution heatmap $H$ for 2D point prediction. 
More architectural details of the decoder are provided in the supplementary material.



\subsection{Training Objectives}
\label{sec:loss}
The type of supervision used directly determines the patterns of the synthesized attention map $A_s$ and the final heatmap $H$, making it a crucial component of our method. 

For the Q-Synth Module, we use part-level segmentation masks $M_p \in \mathbb{R}^{P_h \times P_w}$, downsampled from the original segmentation masks $M \in \mathbb{R}^{H \times W}$, as ground truth supervision. 
We adopt the Binary Cross-Entropy (BCE) loss to guide learning. 
Intuitively, the goal is to synthesize a query $q_s$ that exhibits the highest similarity to the visual features of the target region in $K_v$. 
Thus, we formulate this as a \emph{per-patch classification problem} using BCE loss:
\begin{equation}
    \mathcal{L}_{\mathrm{bce}} = \mathrm{BCE}\!\left(q_s K_v^{\top},\, M_p\right),
    \label{bce_loss}
\end{equation}
where $q_s K_v^{\top}$ represents the predicted per-patch similarity scores. 
In this way, the Q-Synth Module is encouraged not only to extract relevant textual information from the text features but also to refine and align its latent embeddings with the corresponding visual features. 
Unlike conventional attention maps, our synthesized attention patterns tend to produce near-binary activations around the target regions, which facilitates effective visual feature modulation in Sec.~\ref{sec:a2p}.

For the A2P Decoder, we aim for the predicted heatmap to not only cover the target region but also concentrate around the most representative point of the object part. 
To this end, we design a penalty field inspired by the Signed Distance Field (SDF) commonly used in 3D reconstruction. 
In a standard SDF, boundary pixels have zero values, with positive distances outside the ground-truth mask and negative distances inside. 
However, this formulation treats both sides of the boundary symmetrically, which is suboptimal for our objective. 
In our case, predictions located outside the target region should incur large penalties, while those inside the mask are considered correct and should receive much smaller penalties; nevertheless, we still want to encourage the model to focus near the innermost point of the target. 

To achieve this, we define a mapping function $f$ that transforms the original SDF into a modified penalty field. 
Specifically, we apply the softplus function to the original distances and introduce two hyperparameters, $\tau$ and $\gamma$, to control the \textit{steepness} and \textit{asymmetry} of the field inside and outside the mask:
\begin{equation}
\label{eq:fz}
f(x) \;=\; \operatorname{softplus}\!\left(\frac{x}{\tau}\right)
\;+\; \gamma
\begin{cases}
e^{x/\tau}, & x \le 0,\\[3pt]
1, & x > 0~,
\end{cases}
\end{equation}
where $x$ denotes a scalar SDF value. 
Further details and visualizations of the penalty field are provided in the supplementary material.
Finally, our SDF-based loss is computed as a spatial sum of the product between the predicted heatmap distribution and the modified penalty field:
\begin{equation}
    \mathcal{L}_{\mathrm{sdf}} 
    = \sum_{u,v} 
    \operatorname{softmax}\!\big(H(u,v)\big)\; f\!\big(D(u,v)\big),
    \label{sdf_loss}
\end{equation}
where $H$ is the predicted heatmap, $D$ is the Signed Distance Field, and $(u,v)$ index pixel positions.

Our total loss is defined as 
$\mathcal{L}_{\mathrm{total}} = \mathcal{L}_{\mathrm{bce}} + \lambda \mathcal{L}_{\mathrm{sdf}}$, 
and we train the model end-to-end. 
Notably, all original MLLM parameters are frozen during training, enabling us to enhance the pointing ability while fully preserving the models’ pre-trained capabilities.

\section{Experiments}
\label{sec:exp}

\subsection{Settings}
\paragraph{Datasets.}
We include three datasets in our experiments: 
\textbf{(1) PACO~\cite{ramanathan2023paco}} is originally designed for part-level segmentation. 
Each annotation follows the format \textit{$<$object$>$:$<$part$>$} (e.g., \textit{knife:handle}). 
For point grounding, we convert the annotations into textual queries such as ``\textit{Point to the handle of the knife.}'' 
Since the instruction directly specifies the target part, we refer to this as the \textit{direct pointing} task. 
\textbf{(2) InstructPart~\cite{wan2024instructpart}} provides instruction–mask pairs at the part level. 
For example, ``\textit{If I want to open the oven, which part should be utilized?}'' 
Such queries often omit the target part and require reasoning, so we term this the \textit{reasoning pointing} task. 
\textbf{(3) PointArena Point-Bench~\cite{pointarena}} is a benchmark for evaluating multimodal pointing ability across five tasks: Affordance, Spatial, Reasoning, Steerability, and Counting. 
Among them, only the Affordance task focuses on part-level pointing, while the others mainly involve object-level grounding. However, we still include these tasks to evaluate the generalizability of our method beyond part-level scenarios. 
The Counting task, which requires numerical outputs, is excluded as it falls outside our framework’s scope. 
Further dataset details are provided in the supplementary material.

\begin{table*}
  \caption{Main quantitative results on the PACO~\cite{ramanathan2023paco} (direct pointing) and InstructPart~\cite{wan2024instructpart} (reasoning pointing) datasets. 
    Our method consistently outperforms all baselines across both point grounding tasks. 
    Notably, even for the MLLM without any point-grounding ability (First-Gen-MLLM), our approach effectively equips it with this capability and yields significant performance gains.}
  \label{tab:main_tab}
  \centering
    \begin{tabular}{l|cc|cc}
    \toprule
     & \multicolumn{2}{c|}{PACO~\cite{ramanathan2023paco}} & \multicolumn{2}{c}{InstructPart~\cite{wan2024instructpart}} \\
    \midrule
     & Patch Accuracy &  Accuracy &  Patch Accuracy &  Accuracy \\
    \midrule
    \rowcolor{lightgray} \multicolumn{5}{l}{\textbf{\textit{Segmentation-based Models}}} \\
    VLPart~\cite{vlpart}   & 0.419 & 0.381 & 0.008 & 0.008 \\
    X-Decoder~\cite{xdecoder}   & 0.031 & 0.025 & 0.185  & 0.178 \\
    \midrule
    \midrule        
    \rowcolor{lightgray} \multicolumn{5}{l}{\textbf{\textit{MLLM w/ pointing ability}}} \\
    Molmo-7B text pointing~\cite{deitke2025molmo}   & 0.559 & 0.487 & 0.737  & 0.710 \\
    Molmo-7B attention pointing~\cite{kang2025your}   & 0.517 & 0.428 & 0.468 & 0.378 \\
    Molmo-7B \textbf{Ours}   & \textbf{0.603}  & \textbf{0.510} & \textbf{0.900}  & \textbf{0.868} \\
    \midrule
    Qwen2.5-VL-7B text pointing~\cite{bai2025qwen25vltechnicalreport}   & 0.491 & 0.407 & 0.722  & 0.708 \\
    Qwen2.5-VL-7B attention pointing~\cite{kang2025your}   & 0.424 & 0.309 & 0.352  & 0.283 \\
    Qwen2.5-VL-7B \textbf{Ours}   & \textbf{0.610} & \textbf{0.479} & \textbf{0.877}  & \textbf{0.818} \\
    \midrule
    \midrule
    \rowcolor{lightgray} \multicolumn{5}{l}{\textbf{\textit{MLLM w/o pointing ability}}} \\
    LLaVA-1.5-7B text pointing   & 0.085 & 0.068 & 0.040  & 0.033 \\
    LLaVA-1.5-7B attention pointing~\cite{kang2025your}  & 0.230 & 0.183 & 0.227  & 0.194 \\
    LLaVA-1.5-7B \textbf{Ours}  & \textbf{0.544} & \textbf{0.463} & \textbf{0.803}   & \textbf{0.783} \\
    \bottomrule
    \end{tabular}
\end{table*}

\paragraph{Segmentation-based Baselines.}
We evaluate two categories of segmentation-based models: 
\textbf{(1) Open-vocabulary part-level segmentation.} 
We adopt VLPart~\cite{vlpart} as a representative model, since it is trained on the PACO dataset~\cite{ramanathan2023paco} for part-level segmentation. 
However, it lacks reasoning ability—although it accepts free-form text inputs, it is expected to fail on the reasoning pointing task in the InstructPart dataset~\cite{wan2024instructpart}. 
\textbf{(2) Open-vocabulary referring segmentation.} 
We select X-Decoder~\cite{xdecoder} for this category. 
Although it is not specifically trained on part-level data, it can process longer referring expressions and is therefore expected to perform better on reasoning pointing tasks. 
More segmentation-based baselines are included in the supplementary material.
For all segmentation-based models, we extract the innermost point of the largest predicted mask as the final point prediction.

\paragraph{MLLMs.}
We evaluate two \textit{pointing-capable} MLLMs, Molmo-7B~\cite{deitke2025molmo} and Qwen2.5-VL-7B~\cite{bai2025qwen25vltechnicalreport}, which have been trained with point-grounding objectives.
We also include one First-Gen-MLLM, defined as a \textit{non-pointing} MLLM under 10B parameters without point-grounding supervision.
Notably, Molmo~\cite{deitke2025molmo} is specialized for pointing and attains state-of-the-art results among open-source models.

We consider three evaluation modes: 
\textbf{(1) Text pointing}: the MLLM outputs the target coordinates in text form through autoregressive generation.
\textbf{(2) Attention pointing}: we use the native text-to-image attention maps from the selected localization heads~\cite{kang2025your} and identify the peak-attention patch for pointing.
\textbf{(3) Ours}: we apply our Q-Synth and A2P modules to each MLLM, keeping the backbone frozen, and evaluate whether our method can consistently improve point grounding across open-source models.

\paragraph{Metrics.}
Following the PointArena benchmark~\cite{pointarena}, we evaluate using a hit-or-miss criterion, where a prediction is considered correct if the predicted point lies within the ground-truth mask. 
Accuracy is then computed as the ratio of hits to total samples. 
As in the PointArena protocol, when an MLLM outputs multiple points in text form, we take the first predicted point as the final output. 
This setup naturally aligns with our motivation for robotic interaction, where the model generates a single actionable point at each step to guide the robot’s manipulation. 

In addition, attention-based methods, including ours, produce attention maps and locate the target by pointing to the center of the peak attention patch.  Therefore, we can also compute \textit{patch accuracy}, which reflects the localization precision before patch-level quantization errors.
Although our A2P Decoder alleviates this through upsampling, some residual error remains in the final heatmap.
For non-attention-based methods, we convert their point predictions back to the patch coordinates to enable fair comparison under the same metric.

\begin{figure*}[tp]
    \centering
    \includegraphics[width=\linewidth]{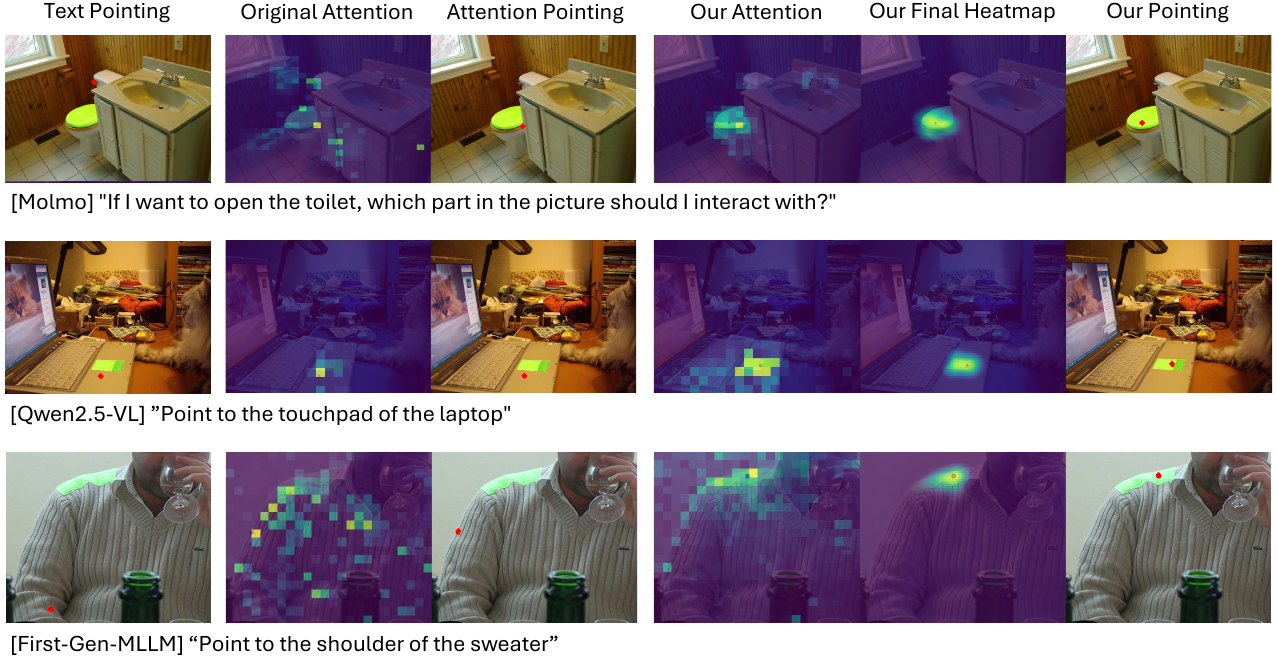}
    \caption{Qualitative results. We compare text pointing, attention pointing, and our proposed method across columns. 
    Each row presents examples from different MLLMs, with the model and text instruction shown below the images. 
    In the visualizations, green masks denote ground-truth regions, and red points indicate the predicted point grounding locations.
    Notably, the sweater-part example in the third row echoes our motivation in Sec.~\ref{sec:intro}, highlighting the relevance of part-level point grounding for tasks such as cloth folding.}
    \label{fig:qualitative}
\end{figure*}

\subsection{Quantitative Results}
We present our main quantitative results in Table~\ref{tab:main_tab}. 
Across both the direct pointing task on the PACO dataset~\cite{ramanathan2023paco} and the reasoning pointing task on the InstructPart dataset~\cite{wan2024instructpart}, our proposed method consistently outperforms all baselines, including the state-of-the-art pointing-specialized MLLM Molmo~\cite{deitke2025molmo}. 

Notably, compared to the attention-pointing baseline, our approach achieves substantial improvements, especially on the InstructPart dataset~\cite{wan2024instructpart}, where the input texts are generally longer and require reasoning. This indicates that our method effectively extracts relevant textual semantics and refines the native attention patterns to yield more accurate and spatially aligned grounding maps (see Sec.~\ref{qualitative} for qualitative comparisons). 
Furthermore, even for a model without any point-grounding capability (First-Gen-MLLM), our method can still leverage its internal attention mechanism to produce competitive pointing results comparable to those of pointing-capable models. 
This demonstrates the generality of our framework: as long as an MLLM incorporates an attention mechanism—which is true for virtually all existing MLLMs—our method can enhance its pointing ability while fully preserving its pre-trained reasoning and language understanding capabilities, since no parameters of the original model are modified.


\subsection{Qualitative Results}
\label{qualitative}
To better illustrate the improved attention behavior of our method, we present qualitative examples in Figure~\ref{fig:qualitative}. 
Our method produces attention maps that more completely cover the target regions, benefiting from the design of the Q-Synth Module and the BCE-based supervision. 
Moreover, the A2P Decoder further refines these patch-level attention maps and generates high-resolution heatmaps that lead to more accurate point predictions. 
Additional visualizations are provided in the supplementary material.


\subsection{Generalizability}
\begin{table*}
  \caption{Generalizability experiment results. We conduct cross-dataset evaluation on the PointArena Point-Bench~\cite{pointarena} after training on PACO~\cite{ramanathan2023paco} (direct pointing). 
  Our method generalizes to unseen tasks and improves over native attention pointing, showing that the proposed components effectively condense textual meaning into grounding-aware queries that align with target visual concepts and produce more meaningful attention patterns across datasets and tasks.
  P-Acc. and Acc. denote Patch Accuracy and Accuracy, respectively.}
  \label{tab:generalize_tab}
  \centering
    \begin{tabular}{l|cc|cc|cc|cc}
    \toprule
     & \multicolumn{2}{c|}{Affordance} & \multicolumn{2}{c|}{Spatial} & \multicolumn{2}{c|}{Reasoning} & \multicolumn{2}{c}{Steerability}\\
    \midrule
     & P-Acc. &  Acc. &  P-Acc. &  Acc. & P-Acc. &  Acc. & P-Acc. &  Acc. \\
    \midrule
    Molmo-7B attention pointing~\cite{kang2025your}   & 0.495 & 0.419 & 0.436 & 0.369 & 0.513 & 0.430 & 0.285 & 0.275 \\
    Molmo-7B \textbf{Ours}   & \textbf{0.793}  & \textbf{0.748} & \textbf{0.554}  & \textbf{0.508} & \textbf{0.653} & \textbf{0.591} & \textbf{0.450} & \textbf{0.410} \\
    \midrule
    Qwen2.5-VL-7B attention pointing~\cite{kang2025your}   & 0.359 & 0.303 & 0.431 & 0.380 & 0.373 & 0.321 & 0.227 & 0.172 \\
    Qwen2.5-VL-7B \textbf{Ours}   & \textbf{0.838} & \textbf{0.828} & \textbf{0.585}  & \textbf{0.544} & \textbf{0.658} & \textbf{0.622} & \textbf{0.430} & \textbf{0.390} \\
    \midrule
    First-Gen-MLLM attention pointing~\cite{kang2025your}  & 0.510 & 0.417 & 0.256  & 0.232 & 0.389 & 0.322 & 0.173 & 0.167 \\
    First-Gen-MLLM \textbf{Ours}  & \textbf{0.672} & \textbf{0.641} & \textbf{0.424} & \textbf{0.384} & \textbf{0.439} & \textbf{0.378} & \textbf{0.405} & \textbf{0.369} \\
    \bottomrule
    \end{tabular}
\end{table*}
We train on the PACO~\cite{ramanathan2023paco} dataset with \emph{direct pointing} supervision and evaluate on the PointArena benchmark~\cite{pointarena} with \emph{reasoning pointing} prompts to assess generalizability of our method. 
The results are shown in Table~\ref{tab:generalize_tab}. 
Among the various tasks in this benchmark, only the Affordance task is similar to PACO, while the others represent unseen tasks during training.
Surprisingly, across all models and tasks, our method consistently improves over the original attention-pointing baseline. 
This transferability demonstrates that our method learns to summarize the overall semantic intent of the input text and align it with the most relevant visual concepts, enabling the model to better associate language and visual evidence even in unseen scenarios. 
In other words, the Q-Synth Module learns to condense textual meaning into a grounding-aware query, while the A2P Decoder translates refined attention into spatially coherent heatmaps, yielding more interpretable attention patterns across different grounding tasks and datasets.


\subsection{Ablation Study}
We conduct ablations on the First-Gen-MLLM using the PACO~\cite{ramanathan2023paco} dataset, as the model lacks inherent pointing capability; hence, any improvement on the pointing task can be directly attributed to our proposed components. More ablation studies are provided in the supplementary material.

\paragraph{Proposed Components.}
\begin{table}
  \caption{Ablation study of the proposed components. 
  Removing the Q-Synth Module leads to the largest drop, indicating that the main limitation lies in suboptimal native attention patterns.
  Therefore, synthesizing a grounding-aware query is crucial for extracting more accurate attention and enhancing the grounding ability of MLLMs.}
  \label{tab:ablation_method}
  \centering
    \begin{tabular}{ccc|c}
    \toprule
     Q-Synth & A2P Decoder & Image $V$ & Accuracy \\
    \midrule
    \checkmark & \checkmark & \checkmark & \textbf{0.463} \\
    \midrule
    \checkmark & \checkmark &  & 0.453 \\
    \midrule
    \checkmark &  &  & 0.429 \\
    \midrule
     & \checkmark & \checkmark & 0.336 \\
    \bottomrule
    \end{tabular}
\end{table}
Table~\ref{tab:ablation_method} presents the ablation results of key components in our framework. 
We first examine the effect of using the image $V$ as input to the A2P Decoder. 
Without these features, the decoder only receives the $k$-channel attention maps concatenated from $k$ localization heads, leading to suboptimal results due to the lack of fine-grained visual information.

Next, we remove the A2P Decoder entirely and use only the low-resolution attention maps for point prediction. 
Following~\cite{kang2025your}, we aggregate attention maps across localization heads via element-wise summation. 
While performance decreases further ($0.429$), it remains noticeably higher than the baseline attention pointing from First-Gen-MLLM ($0.183$) as shown in Table~\ref{tab:main_tab}. 
This indicates that the Q-Synth Module alone can produce more accurate and semantically consistent attention patterns.

Finally, removing the Q-Synth Module and using only the original attention maps as input to the A2P Decoder results in the largest performance drop. 
This confirms that the main limitation lies not merely in the low resolution of the attention maps but in their semantic precision. 
Even with upsampling and spatial refinement, poor attention initialization limits grounding quality. 
Therefore, synthesizing a grounding-aware query via the Q-Synth Module is crucial for enhancing the point grounding ability of an MLLM.

\paragraph{Number of Selected heads.}
\begin{table}
  \caption{Ablation on the number of selected localization heads.
  Increasing $k$ improves performance due to richer feature diversity for the A2P Decoder.} 
  \label{tab:ablation_k}
  \centering
    \begin{tabular}{lccc}
    \toprule
    & $k$ = 1 & $k$ = 3 & $k$ = 5 (Ours) \\
    \midrule
    Accuracy & 0.427 & 0.451 & \textbf{0.463} \\
    \bottomrule
    \end{tabular}
\end{table}
In Table~\ref{tab:ablation_k}, we present the performance when using different numbers of selected localization heads. 
We observe that increasing $k$ generally improves pointing accuracy. 
This trend differs from the finding in~\cite{kang2025your}, where $k{=}3$ yielded the best average performance. 
We attribute this difference to our A2P Decoder, which leverages image $V$ features, and benefits from a larger set of heads that provides a richer and more diverse feature basis for refinement. 
When $k$ exceeds 5, additional heads show weaker localization and less distinctive attention patterns, while computation also increases. Therefore, we do not consider larger values of $k$ in our experiments.
\section{Conclusion}
\label{sec:conclusion}


In this work, we introduced a general framework that enhances Multimodal Large Language Models (MLLMs) with accurate part-level point grounding while preserving their original pre-trained capabilities.
Our method refines native attention patterns through the Query Synthesis Module, and the Attention-to-Point Decoder produces high-resolution, point-centric heatmaps guided by an SDF-based penalty field, enabling precise and interpretable visual grounding.
Experiments across diverse MLLMs and datasets demonstrate that our framework universally strengthens point grounding ability without retraining the base models, offering a scalable way to adapt future MLLMs for fine-grained grounding tasks.
{
    \small
    \bibliographystyle{ieeenat_fullname}
    \bibliography{main}

@inproceedings{kang2025your,
  title={Your large vision-language model only needs a few attention heads for visual grounding},
  author={Kang, Seil and Kim, Jinyeong and Kim, Junhyeok and Hwang, Seong Jae},
  booktitle={Proceedings of the Computer Vision and Pattern Recognition Conference},
  pages={9339--9350},
  year={2025}
}

@inproceedings{ramanathan2023paco,
  title={Paco: Parts and attributes of common objects},
  author={Ramanathan, Vignesh and Kalia, Anmol and Petrovic, Vladan and Wen, Yi and Zheng, Baixue and Guo, Baishan and Wang, Rui and Marquez, Aaron and Kovvuri, Rama and Kadian, Abhishek and others},
  booktitle={Proceedings of the IEEE/CVF Conference on Computer Vision and Pattern Recognition},
  pages={7141--7151},
  year={2023}
}

@inproceedings{
  wan2024instructpart,
  title={InstructPart: Task-Oriented Part Segmentation with Instruction Reasoning},
  author={Wan, Zifu and  Xie, Yaqi and Zhang, Ce and Lin, Zhiqiu and Wang, Zihan and Stepputtis, Simon and Ramanan, Deva and Sycara, Katia},
  booktitle={The 63rd Annual Meeting of the Association for Computational Linguistics},
  year={2025},
  url={https://openreview.net/forum?id=IMEr4XgJSZ}
}

@misc{pointarena,
      title={PointArena: Probing Multimodal Grounding Through Language-Guided Pointing}, 
      author={Long Cheng and Jiafei Duan and Yi Ru Wang and Haoquan Fang and Boyang Li and Yushan Huang and Elvis Wang and Ainaz Eftekhar and Jason Lee and Wentao Yuan and Rose Hendrix and Noah A. Smith and Fei Xia and Dieter Fox and Ranjay Krishna},
      year={2025},
      eprint={2505.09990},
      archivePrefix={arXiv},
      primaryClass={cs.CV},
      url={https://arxiv.org/abs/2505.09990}, 
}

@inproceedings{vlpart,
  title={Going denser with open-vocabulary part segmentation},
  author={Sun, Peize and Chen, Shoufa and Zhu, Chenchen and Xiao, Fanyi and Luo, Ping and Xie, Saining and Yan, Zhicheng},
  booktitle={Proceedings of the IEEE/CVF International Conference on Computer Vision},
  pages={15453--15465},
  year={2023}
}

@inproceedings{xdecoder,
  title={Generalized decoding for pixel, image, and language},
  author={Zou, Xueyan and Dou, Zi-Yi and Yang, Jianwei and Gan, Zhe and Li, Linjie and Li, Chunyuan and Dai, Xiyang and Behl, Harkirat and Wang, Jianfeng and Yuan, Lu and others},
  booktitle={Proceedings of the IEEE/CVF conference on computer vision and pattern recognition},
  pages={15116--15127},
  year={2023}
}

@inproceedings{lai2024lisa,
  title={Lisa: Reasoning segmentation via large language model},
  author={Lai, Xin and Tian, Zhuotao and Chen, Yukang and Li, Yanwei and Yuan, Yuhui and Liu, Shu and Jia, Jiaya},
  booktitle={Proceedings of the IEEE/CVF Conference on Computer Vision and Pattern Recognition},
  pages={9579--9589},
  year={2024}
}

@inproceedings{deitke2025molmo,
  title={Molmo and pixmo: Open weights and open data for state-of-the-art vision-language models},
  author={Deitke, Matt and Clark, Christopher and Lee, Sangho and Tripathi, Rohun and Yang, Yue and Park, Jae Sung and Salehi, Mohammadreza and Muennighoff, Niklas and Lo, Kyle and Soldaini, Luca and others},
  booktitle={Proceedings of the Computer Vision and Pattern Recognition Conference},
  pages={91--104},
  year={2025}
}

@misc{bai2025qwen25vltechnicalreport,
      title={Qwen2.5-VL Technical Report}, 
      author={Shuai Bai and Keqin Chen and Xuejing Liu and Jialin Wang and Wenbin Ge and Sibo Song and Kai Dang and Peng Wang and Shijie Wang and Jun Tang and Humen Zhong and Yuanzhi Zhu and Mingkun Yang and Zhaohai Li and Jianqiang Wan and Pengfei Wang and Wei Ding and Zheren Fu and Yiheng Xu and Jiabo Ye and Xi Zhang and Tianbao Xie and Zesen Cheng and Hang Zhang and Zhibo Yang and Haiyang Xu and Junyang Lin},
      year={2025},
      eprint={2502.13923},
      archivePrefix={arXiv},
      primaryClass={cs.CV},
      url={https://arxiv.org/abs/2502.13923}, 
}

@article{motoda2025suctionprompt,
  title={SuctionPrompt: Visual-Assisted Robotic Picking with a Suction Cup Using Vision-Language Models and Facile Hardware Design},
  author={Motoda, Tomohiro and Kitamura, Takahide and Hanai, Ryo and Domae, Yukiyasu},
  journal={Journal of Robotics and Mechatronics},
  volume={37},
  number={2},
  pages={374--386},
  year={2025},
  publisher={Fuji Technology Press Ltd.}
}

@article{huang2024rekep,
  title = {ReKep: Spatio-Temporal Reasoning of Relational Keypoint Constraints for Robotic Manipulation},
  author = {Huang, Wenlong and Wang, Chen and Li, Yunzhu and Zhang, Ruohan and Fei-Fei, Li},
  journal = {arXiv preprint arXiv:2409.01652},
  year = {2024}
}

@article{yuan2024robopoint,
  title={RoboPoint: A Vision-Language Model for Spatial Affordance Prediction for Robotics},
  author={Yuan, Wentao and Duan, Jiafei and Blukis, Valts and Pumacay, Wilbert and Krishna, Ranjay and Murali, Adithyavairavan and Mousavian, Arsalan and Fox, Dieter},
  journal={arXiv preprint arXiv:2406.10721},
  year={2024}
}

@article{achiam2023gpt,
  title={Gpt-4 technical report},
  author={Achiam, Josh and Adler, Steven and Agarwal, Sandhini and Ahmad, Lama and Akkaya, Ilge and Aleman, Florencia Leoni and Almeida, Diogo and Altenschmidt, Janko and Altman, Sam and Anadkat, Shyamal and others},
  journal={arXiv preprint arXiv:2303.08774},
  year={2023}
}

@article{team2025gemini,
  title={Gemini robotics: Bringing ai into the physical world},
  author={Team, Gemini Robotics and Abeyruwan, Saminda and Ainslie, Joshua and Alayrac, Jean-Baptiste and Arenas, Montserrat Gonzalez and Armstrong, Travis and Balakrishna, Ashwin and Baruch, Robert and Bauza, Maria and Blokzijl, Michiel and others},
  journal={arXiv preprint arXiv:2503.20020},
  year={2025}
}

@article{chen2023shikra,
  title={Shikra: Unleashing multimodal llm's referential dialogue magic},
  author={Chen, Keqin and Zhang, Zhao and Zeng, Weili and Zhang, Richong and Zhu, Feng and Zhao, Rui},
  journal={arXiv preprint arXiv:2306.15195},
  year={2023}
}

@article{peng2023kosmos,
  title={Kosmos-2: Grounding multimodal large language models to the world},
  author={Peng, Zhiliang and Wang, Wenhui and Dong, Li and Hao, Yaru and Huang, Shaohan and Ma, Shuming and Wei, Furu},
  journal={arXiv preprint arXiv:2306.14824},
  year={2023}
}

@article{you2023ferret,
  title={Ferret: Refer and ground anything anywhere at any granularity},
  author={You, Haoxuan and Zhang, Haotian and Gan, Zhe and Du, Xianzhi and Zhang, Bowen and Wang, Zirui and Cao, Liangliang and Chang, Shih-Fu and Yang, Yinfei},
  journal={arXiv preprint arXiv:2310.07704},
  year={2023}
}

@inproceedings{pi2024perceptiongpt,
  title={Perceptiongpt: Effectively fusing visual perception into llm},
  author={Pi, Renjie and Yao, Lewei and Gao, Jiahui and Zhang, Jipeng and Zhang, Tong},
  booktitle={Proceedings of the IEEE/CVF conference on computer vision and pattern recognition},
  pages={27124--27133},
  year={2024}
}

@inproceedings{rasheed2024glamm,
  title={Glamm: Pixel grounding large multimodal model},
  author={Rasheed, Hanoona and Maaz, Muhammad and Shaji, Sahal and Shaker, Abdelrahman and Khan, Salman and Cholakkal, Hisham and Anwer, Rao M and Xing, Eric and Yang, Ming-Hsuan and Khan, Fahad S},
  booktitle={Proceedings of the IEEE/CVF Conference on Computer Vision and Pattern Recognition},
  pages={13009--13018},
  year={2024}
}

@inproceedings{ren2024pixellm,
  title={Pixellm: Pixel reasoning with large multimodal model},
  author={Ren, Zhongwei and Huang, Zhicheng and Wei, Yunchao and Zhao, Yao and Fu, Dongmei and Feng, Jiashi and Jin, Xiaojie},
  booktitle={Proceedings of the IEEE/CVF Conference on Computer Vision and Pattern Recognition},
  pages={26374--26383},
  year={2024}
}

@article{wei2024lasagna,
  title={Lasagna: Language-based segmentation assistant for complex queries},
  author={Wei, Cong and Tan, Haoxian and Zhong, Yujie and Yang, Yujiu and Ma, Lin},
  journal={arXiv preprint arXiv:2404.08506},
  year={2024}
}

@inproceedings{wu2025flmm,
  title={F-lmm: Grounding frozen large multimodal models},
  author={Wu, Size and Jin, Sheng and Zhang, Wenwei and Xu, Lumin and Liu, Wentao and Li, Wei and Loy, Chen Change},
  booktitle={Proceedings of the Computer Vision and Pattern Recognition Conference},
  pages={24710--24721},
  year={2025}
}

@misc{llava1.5,
      title={Improved Baselines with Visual Instruction Tuning}, 
      author={Haotian Liu and Chunyuan Li and Yuheng Li and Yong Jae Lee},
      year={2024},
      eprint={2310.03744},
      archivePrefix={arXiv},
      primaryClass={cs.CV},
      url={https://arxiv.org/abs/2310.03744}, 
}

@misc{lin2015microsoftcococommonobjects,
      title={Microsoft COCO: Common Objects in Context}, 
      author={Tsung-Yi Lin and Michael Maire and Serge Belongie and Lubomir Bourdev and Ross Girshick and James Hays and Pietro Perona and Deva Ramanan and C. Lawrence Zitnick and Piotr Dollár},
      year={2015},
      eprint={1405.0312},
      archivePrefix={arXiv},
      primaryClass={cs.CV},
      url={https://arxiv.org/abs/1405.0312}, 
}

@inproceedings{
  zhang2025mllms,
  title={{MLLM}s Know Where to Look: Training-free Perception of Small Visual Details with Multimodal {LLM}s},
  author={Jiarui Zhang and Mahyar Khayatkhoei and Prateek Chhikara and Filip Ilievski},
  booktitle={The Thirteenth International Conference on Learning Representations},
  year={2025},
  url={https://arxiv.org/abs/2502.17422}
}

@article{fangandliu2024moka,
      title={MOKA: Open-World Robotic Manipulation through Mark-Based Visual Prompting},
      author={Kuan Fang and Fangchen Liu and Pieter Abbeel and Sergey Levine},
      journal={Robotics: Science and Systems (RSS)},
      year={2024}
  }

@inproceedings{manuelli2019kpam,
  title={kpam: Keypoint affordances for category-level robotic manipulation},
  author={Manuelli, Lucas and Gao, Wei and Florence, Peter and Tedrake, Russ},
  booktitle={The International Symposium of Robotics Research},
  pages={132--157},
  year={2019},
  organization={Springer}
}

@inproceedings{qin2020keto,
  title={Keto: Learning keypoint representations for tool manipulation},
  author={Qin, Zengyi and Fang, Kuan and Zhu, Yuke and Fei-Fei, Li and Savarese, Silvio},
  booktitle={2020 IEEE International Conference on Robotics and Automation (ICRA)},
  pages={7278--7285},
  year={2020},
  organization={IEEE}
}

@article{driess2023palm,
  title={Palm-e: An embodied multimodal language model},
  author={Driess, Danny and Xia, Fei and Sajjadi, Mehdi SM and Lynch, Corey and Chowdhery, Aakanksha and Ichter, Brian and Wahid, Ayzaan and Tompson, Jonathan and Vuong, Quan and Yu, Tianhe and others},
  journal={arXiv preprint arXiv:2303.03378},
  year={2023}
}

@inproceedings{wu2024mldt,
  title={Mldt: Multi-level decomposition for complex long-horizon robotic task planning with open-source large language model},
  author={Wu, Yike and Zhang, Jiatao and Hu, Nan and Tang, Lanling and Qi, Guilin and Shao, Jun and Ren, Jie and Song, Wei},
  booktitle={International Conference on Database Systems for Advanced Applications},
  pages={251--267},
  year={2024},
  organization={Springer}
}

@inproceedings{yang2025guiding,
  title={Guiding long-horizon task and motion planning with vision language models},
  author={Yang, Zhutian and Garrett, Caelan and Fox, Dieter and Lozano-P{\'e}rez, Tom{\'a}s and Kaelbling, Leslie Pack},
  booktitle={2025 IEEE International Conference on Robotics and Automation (ICRA)},
  pages={16847--16853},
  year={2025},
  organization={IEEE}
}

@inproceedings{li2025towards,
  title={Towards Long-Horizon Vision-Language-Action System: Reasoning, Acting and Memory},
  author={Li, Daixun and Zhang, Yusi and Cao, Mingxiang and Liu, Donglai and Xie, Weiying and Hui, Tianlin and Lin, Lunkai and Xie, Zhiqiang and Li, Yunsong},
  booktitle={Proceedings of the IEEE/CVF International Conference on Computer Vision},
  pages={6839--6848},
  year={2025}
}
}

\clearpage
\appendix
\gdef\thesection{\Alph{section}} 
\setcounter{page}{1}
\maketitlesupplementary


\begin{figure*}[t!]
    \centering
    \includegraphics[width=\linewidth]{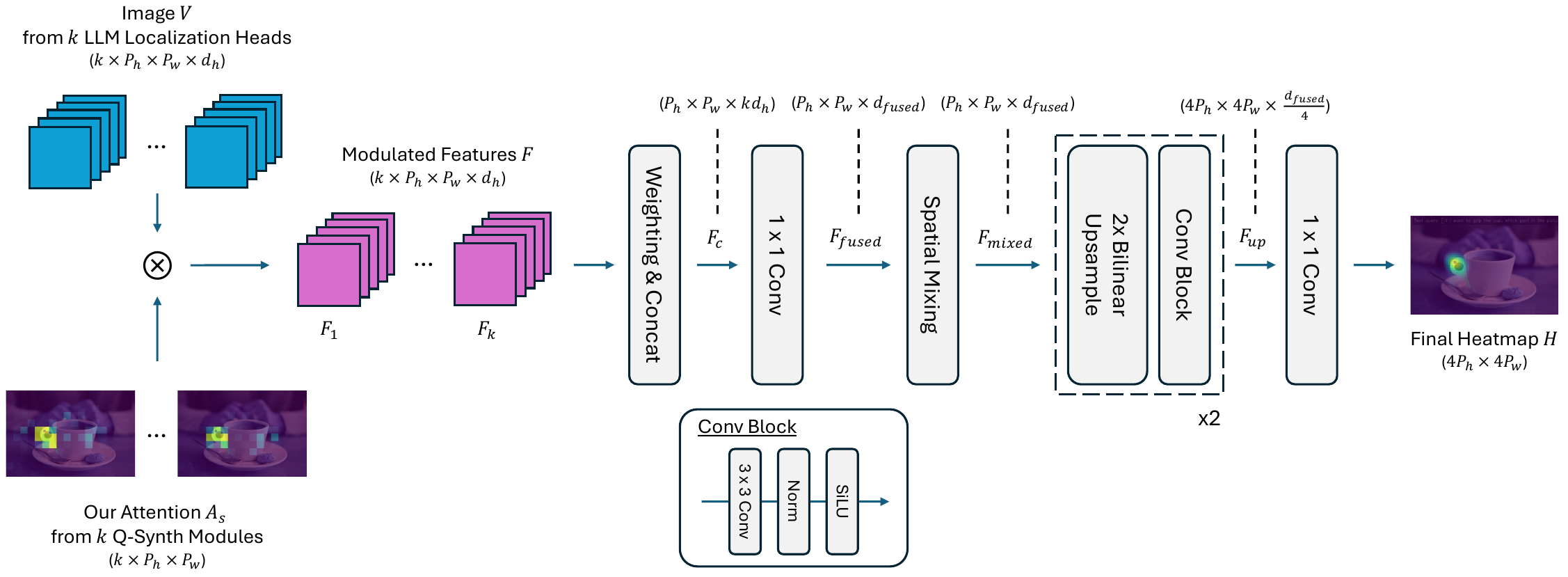}
    \caption{Architecture of the proposed Attention-to-Point (A2P) Decoder. The A2P Decoder fuses $k$ attention maps with image $V$ features and upsamples them into a high-resolution, point-centric heatmap for 2D point grounding.
    The Spatial Mixing consists of four Conv Blocks with skip connections.}
    \label{fig:supp_a2p}
\end{figure*}

\section{Overview}
This supplementary material provides additional details and results that complement the main manuscript. In Sec.~\ref{supp_a2p}, we describe the architectural details of the proposed Attention-to-Point (A2P) Decoder. In Sec.~\ref{supp_sdf}, we present plots and visualizations of the proposed SDF mapping function and the resulting penalty fields to further illustrate the intuition behind our design. In Sec.~\ref{supp_dataset}, we provide statistics and preprocessing steps for the three datasets used in our experiments. 
In Sec.~\ref{supp_implementation}, we outline implementation details. 
Sec.~\ref{supp_baselines} and Sec.~\ref{supp_qualitative} present additional quantitative and qualitative results, respectively. 
More ablation studies can be found in Sec.~\ref{supp_ablation}. Besides, we provide information about the latency and model size in Sec.~\ref{supp_lantency}.
Finally, we discuss the limitations of our method and potential future directions in Sec.~\ref{supp_limitation}.

\section{Attention-to-Point (A2P) Decoder}
\label{supp_a2p}
Figure~\ref{fig:supp_a2p} presents the architecture of the proposed Attention-to-Point (A2P) Decoder. 
Given $k$ synthesized attention maps and image $V$ features from $k$ Q-Synth Modules and LLM localization heads, the A2P Decoder aims to fuse these signals and produce a single high-resolution, point-centric heatmap for 2D point prediction.
Specifically, each attention map first modulates its corresponding image-value feature to obtain a spatially weighted feature map $F_i \in \mathbb{R}^{P_h \times P_w \times d_h}$ for $i \in {1,\dots,k}$. An MLP is then used to learn per-head importance weights over the set ${F_i}$, and the weighted features are concatenated along the channel dimension to form $F_{c} \in \mathbb{R}^{P_h \times P_w \times kd_h}$.
Consquently, a $1\times1$ convolution is applied to fuse information across the $k$ heads, producing $F_\mathrm{fused} \in \mathbb{R}^{P_h \times P_w \times d_{\mathrm{fused}}}$. 
In our experiments, we use $k=5$, $d_h=128$, and $d_{\mathrm{fused}}=256$.
Next, spatial mixing is performed using four convolutional blocks with skip connections, yielding $F_{\mathrm{mixed}}$.
Each block consists of a $3\times3$ convolution, a GroupNorm layer, and a SiLU activation.
To obtain a high-resolution output, we apply two stages of $2\times$ bilinear interpolation, each followed by a convolutional block, resulting in the upsampled feature $F_{\mathrm{up}} \in \mathbb{R}^{4P_h \times 4P_w \times \frac{d_{\mathrm{fused}}}{4}}$.
Finally, a $1\times1$ convolution compresses the channels into a single output map, generating the final heatmap $H$, which has a spatial resolution four times larger than that of the original attention maps.

\section{SDF-based Penalty Field}
\label{supp_sdf}
\begin{figure*}[tp]
    \centering
    \includegraphics[width=0.98\linewidth]{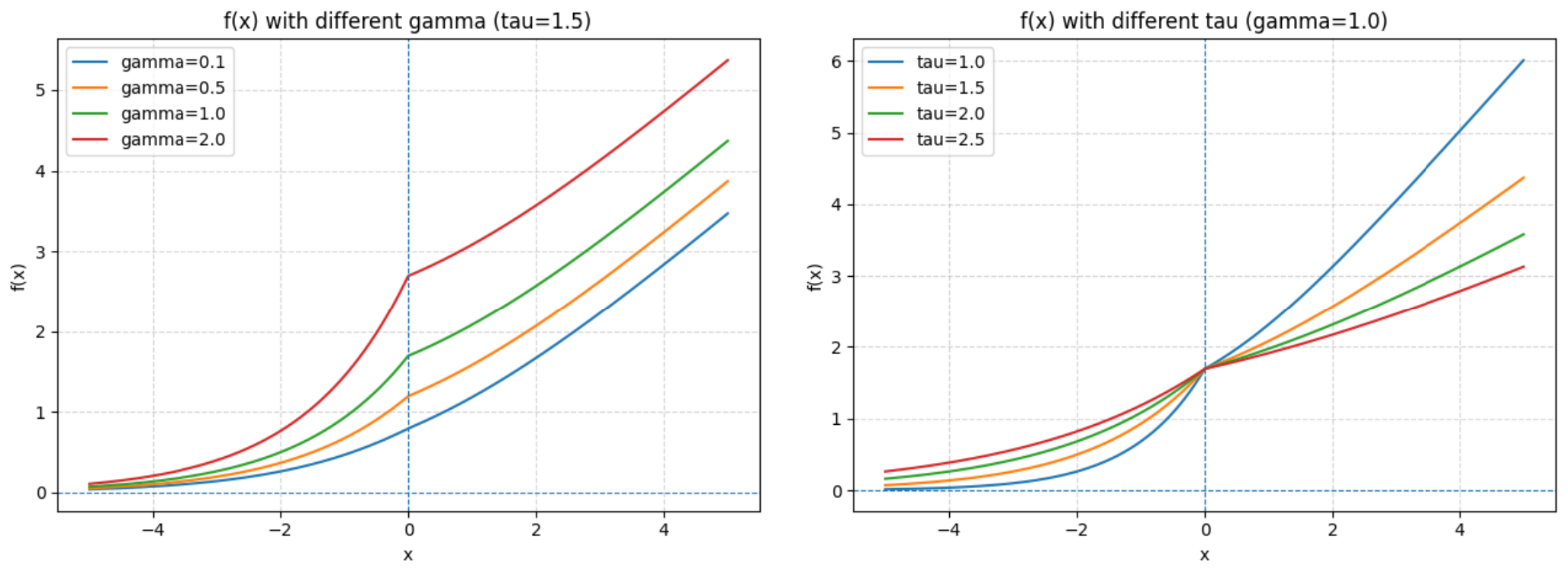}
    \caption{The proposed SDF mapping function $f$ under different $(\tau, \gamma)$ settings. Here, $x$ denotes the original scalar SDF values. In the left plot, $\gamma$ controls the asymmetry of the mapped values inside versus outside the mask (i.e., for $x<0$ vs.\ $x>0$). In the right plot, varying $\tau$ adjusts the overall steepness of the penalty field. 
    The intuition behind this design is that predictions outside the target region should incur large penalties ($x>0$), while predictions inside the mask should receive much smaller penalties ($x<0$). At the same time, the function encourages higher confidence near the innermost point of the target by assigning progressively smaller penalties as $x$ decreases.
    We use $\tau=1.5$ and $\gamma=1$ in our experiments.}
    \label{fig:supp_sdf}
\end{figure*}
\begin{figure}[tp]
    \centering
    \includegraphics[width=0.95\linewidth]{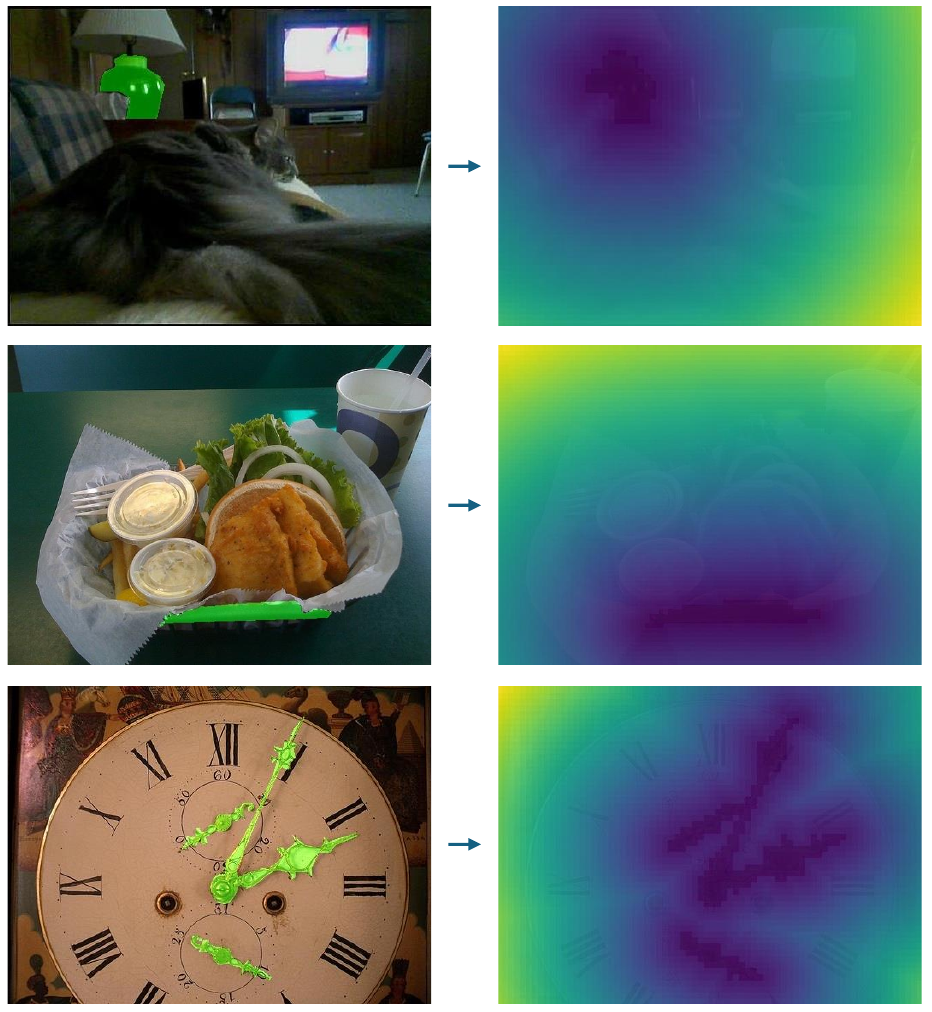}
    \caption{Examples of the original ground-truth masks and the corresponding SDF-based penalty fields. Cooler colors indicate smaller values (i.e., lower penalties). Notably, the third row illustrates a case with multiple object-part instances, where the proposed mapping function still produces a meaningful and well-structured penalty field for supervision.}
    \label{fig:supp_sdf_example}
\end{figure}
To supervise the prediction of point-centric heatmaps, we introduce an SDF-based penalty field tailored for our point grounding task. Specifically, each ground-truth mask is first converted into a Signed Distance Field (SDF). We then design a mapping function $f$ that transforms the raw SDF into a penalty field with properties better aligned to point-level supervision. As described in the main manuscript, the mapping function is defined as
\begin{equation}
\label{eq:fz}
f(x) \;=\; \operatorname{softplus}\!\left(\frac{x}{\tau}\right)
\;+\; \gamma
\begin{cases}
e^{x/\tau}, & x \le 0,\\[3pt]
1, & x > 0~,
\end{cases}
\end{equation}
where $x$ denotes a scalar SDF value, and $\tau$ and $\gamma$ are hyperparameters controlling the steepness and asymmetry of the penalty inside and outside the mask region.

To visualize the effect of different hyperparameter choices, Figure~\ref{fig:supp_sdf} shows plots of the mapping function under varying $(\tau, \gamma)$ values. In our experiments, we use $\tau=1.5$ and $\gamma=1$, which we find to provide the most stable training behavior. Additionally, Figure~\ref{fig:supp_sdf_example} presents several examples comparing the original binary ground-truth masks with their corresponding transformed penalty fields.

\section{Dataset Statistics}
\label{supp_dataset}
\begin{figure*}[tp]
    \centering
    \includegraphics[width=\linewidth]{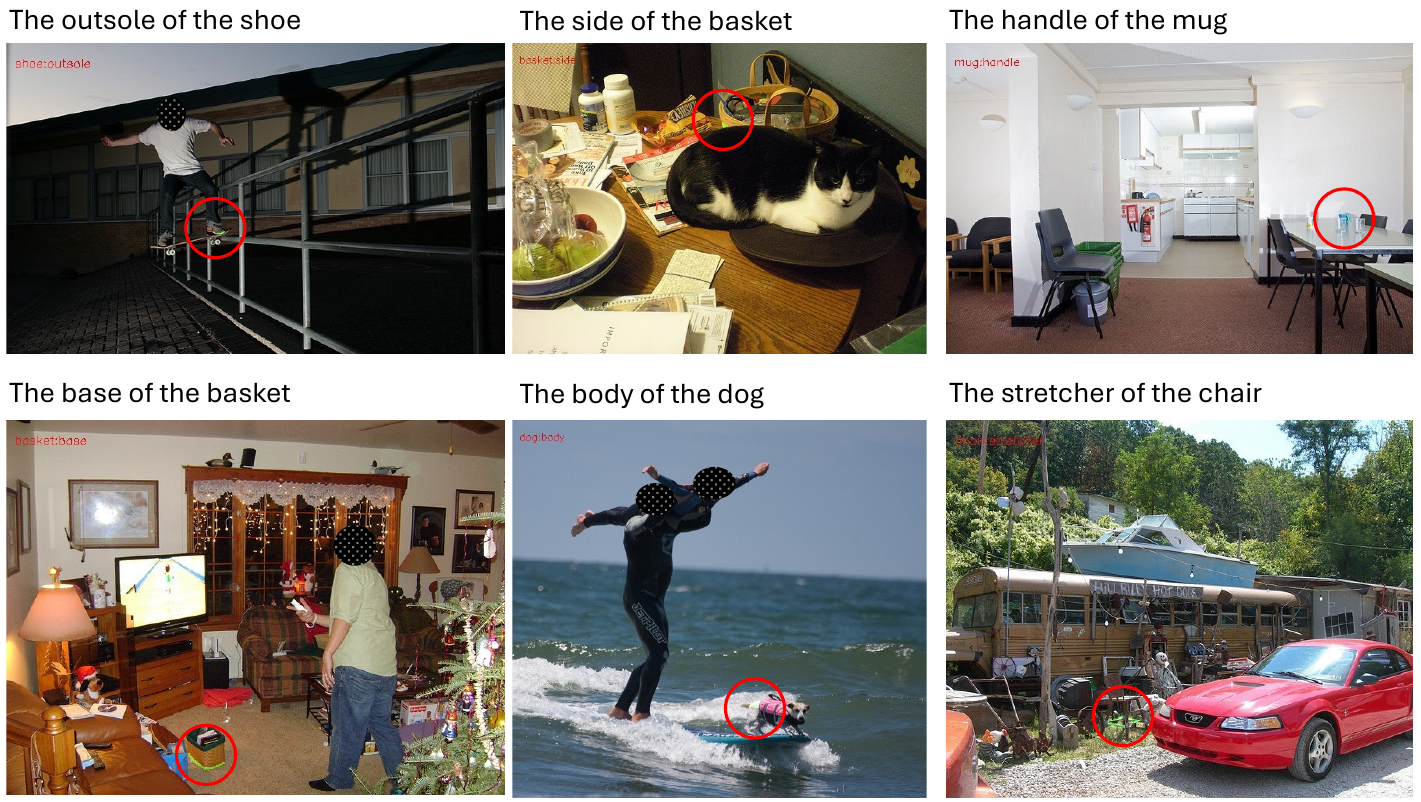}
    \caption{Examples of filtered samples from the PACO~\cite{ramanathan2023paco} dataset due to extremely small part areas. These cases correspond to parts that appear tiny because of heavy occlusion, distant camera viewpoints, or the inherent characteristics of certain categories.
    Red circles highlight the regions of interest, and the green masks within them indicate the tiny ground-truth annotations.}
    \label{fig:supp_paco_filter}
\end{figure*}
For the \textbf{PACO~\cite{ramanathan2023paco} dataset}, we use 132,442 part-level training samples and 28,318 testing samples encompassing 422 object parts in total, after applying several preprocessing steps. First, in the original dataset, different object-part instances within the same image are treated as separate annotations; we merge these annotations into a single combined mask per object part for that image. For images containing multiple instances of the same object part, we consider a point prediction correct if it falls within any of the corresponding masks. This evaluation protocol is applied consistently across all baseline methods and our method.
Next, we filter out samples whose ground-truth segmentation masks occupy less than $0.1\%$ of the image area. Although these small regions are correctly annotated, they typically correspond to extremely tiny parts that are not meaningful for point grounding, either due to heavy occlusion, distant camera viewpoints, or the inherent small size of certain object-part categories. Examples of such filtered cases are shown in Figure~\ref{fig:supp_paco_filter}.
For the \textbf{InstructPart~\cite{wan2024instructpart} dataset}, we use 1,800 training samples and 600 testing samples, and directly adopt the original instruction–mask pairs for our experiments. For the \textbf{PointArena Point-Bench~\cite{pointarena}}, we evaluate on four tasks—Affordance, Spatial, Reasoning, and Steerability—which contain 198, 195, 193, and 200 testing samples, respectively.
Due to GPU memory constraints, we downsample the large input images so that their height and width do not exceed 1,500 pixels.
All baseline methods and our approach are evaluated under the same resolution constraints, ensuring a fair comparison.
\section{Implementation Details}
\label{supp_implementation}
\begin{figure*}[tp]
    \centering
    \includegraphics[width=0.88\linewidth]{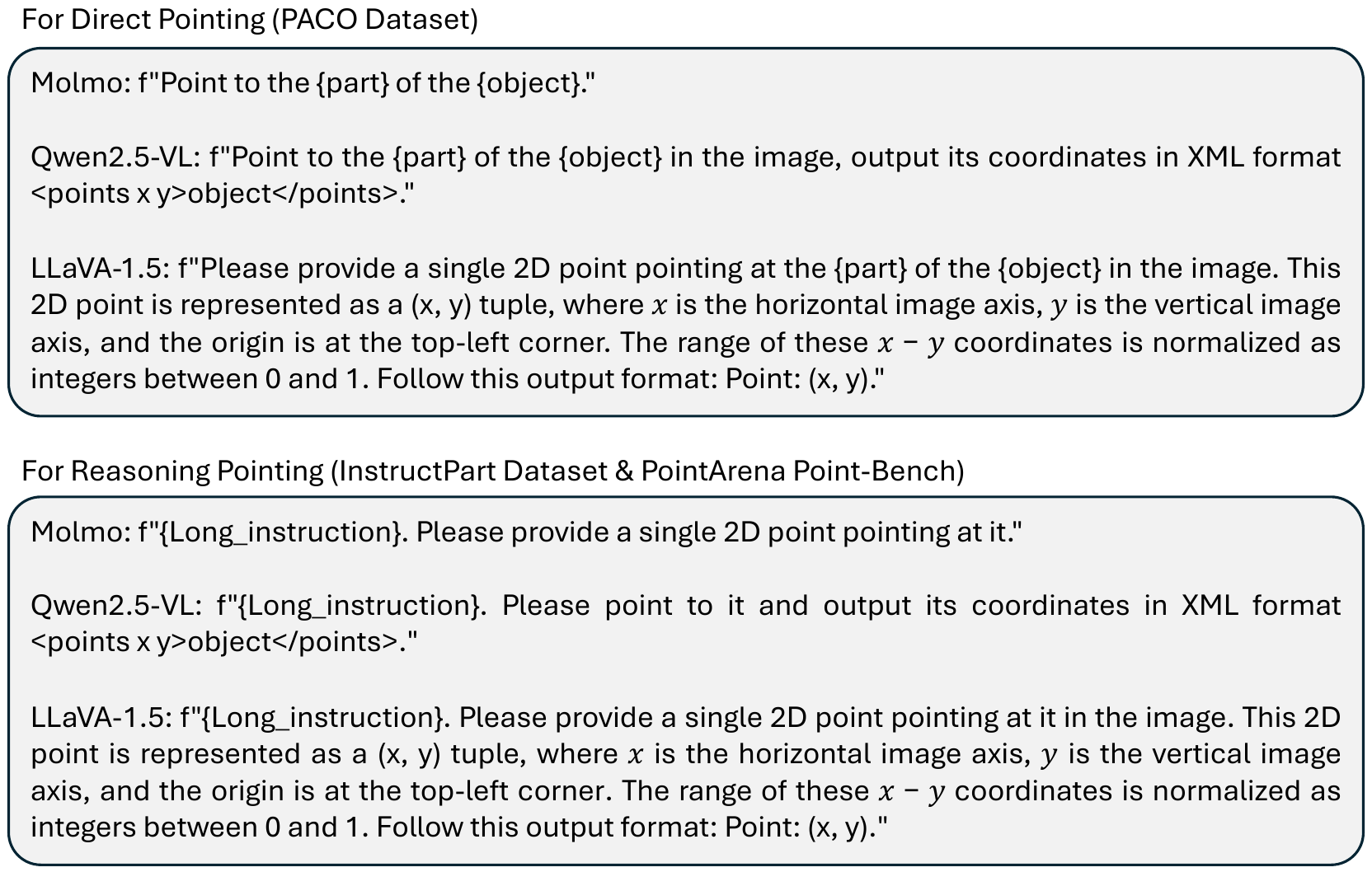}
    \caption{MLLM text-pointing prompts. The Long\_instruction used in the reasoning pointing task includes queries that require contextual understanding. For example: ``If I want to pick up the knife, which part in the picture can be used?''}
    \label{fig:prompt}
\end{figure*}

We provide the additional hyperparameters used in our experiments as follows. For the Query Synthesis (Q-Synth) Module, we initialize $N=4$ learnable latent queries and set the number of cross-attention layers to $T=3$. For the total training loss, we set the weight of the SDF-based loss to $\lambda = 0.001$ to balance its scale relative to the BCE loss. We use AdamW as the optimizer with a learning rate of $1 \times 10^{-4}$ and a weight decay of 0.01, employ a batch size of 4, and train the model for 30 epochs. All experiments can be run on a single NVIDIA A6000 GPU with 48 GB of memory; in practice, we use four A6000 GPUs for data parallelism.

In addition, the First-Gen-MLLM used in our experiments is LLaVA-1.5-7B~\cite{llava1.5}, although in principle any MLLM that does not natively possess pointing ability can be used. The prompts employed for each MLLM are shown in Figure~\ref{fig:prompt}. Note that the prompts differ slightly across datasets due to the nature of the tasks, namely direct pointing versus reasoning pointing.

\section{More Baseline Results}
\label{supp_baselines}
\begin{table*}
  \caption{More baseline results on the PACO~\cite{ramanathan2023paco} (direct pointing) and InstructPart~\cite{wan2024instructpart} (reasoning pointing) datasets. 
    Our method consistently outperforms all baselines across both point grounding tasks.
    Notably, even for the MLLM without any point-grounding ability (First-Gen-MLLM), our approach effectively equips it with this capability and yields significant performance gains.}
  \label{tab:supp_main_tab}
  \centering
    \begin{tabular}{l|cc|cc}
    \toprule
     & \multicolumn{2}{c|}{PACO~\cite{ramanathan2023paco}} & \multicolumn{2}{c}{InstructPart~\cite{wan2024instructpart}} \\
    \midrule
     & Patch Accuracy &  Accuracy &  Patch Accuracy &  Accuracy \\
    \midrule
    \rowcolor{lightgray} \multicolumn{5}{l}{\textbf{\textit{Segmentation-based Models}}} \\
    VLPart~\cite{vlpart}   & 0.419 & 0.381 & 0.008 & 0.008 \\
    X-Decoder~\cite{xdecoder}   & 0.031 & 0.025 & 0.185  & 0.178 \\
    LISA-7B~\cite{lai2024lisa}   & 0.345 & 0.290 & 0.480  & 0.464 \\
    GLaMM-7B~\cite{rasheed2024glamm}   & 0.540 & 0.456 & 0.412  & 0.400 \\
    \midrule
    \midrule        
    \rowcolor{lightgray} \multicolumn{5}{l}{\textbf{\textit{MLLM w/ pointing ability}}} \\
    Molmo-7B text pointing~\cite{deitke2025molmo}   & 0.559 & 0.487 & 0.737  & 0.710 \\
    Molmo-7B attention pointing~\cite{kang2025your}   & 0.517 & 0.428 & 0.468 & 0.378 \\
    Molmo-7B \textbf{Ours}   & \textbf{0.603}  & \textbf{0.510} & \textbf{0.900}  & \textbf{0.868} \\
    \midrule
    Qwen2.5-VL-7B text pointing~\cite{bai2025qwen25vltechnicalreport}   & 0.491 & 0.407 & 0.722  & 0.708 \\
    Qwen2.5-VL-7B attention pointing~\cite{kang2025your}   & 0.424 & 0.309 & 0.352  & 0.283 \\
    Qwen2.5-VL-7B \textbf{Ours}   & \textbf{0.610} & \textbf{0.479} & \textbf{0.877}  & \textbf{0.818} \\
    \midrule
    \midrule
    \rowcolor{lightgray} \multicolumn{5}{l}{\textbf{\textit{MLLM w/o pointing ability}}} \\
    First-Gen-MLLM text pointing   & 0.085 & 0.068 & 0.040  & 0.033 \\
    First-Gen-MLLM attention pointing~\cite{kang2025your}  & 0.230 & 0.183 & 0.227  & 0.194 \\
    First-Gen-MLLM \textbf{Ours}  & \textbf{0.544} & \textbf{0.463} & \textbf{0.803}   & \textbf{0.783} \\
    \bottomrule
    \end{tabular}
\end{table*}
We provide another type of segmentation-based baseline, Fine-tuned MLLM-based reasoning segmentation, in Table~\ref{tab:supp_main_tab}. Specifically, we include LISA~\cite{lai2024lisa} and GLaMM~\cite{rasheed2024glamm}, two representative reasoning-segmentation models. By fine-tuning MLLMs to output a special token \texttt{[SEG]} and decode segmentation masks from it, these methods can handle more complex textual instructions that require reasoning. However, such fine-tuning often degrades the original MLLMs’ reasoning or VQA capabilities, as it forces the model to generate special-purpose tokens that deviate from natural conversational outputs. In contrast, our method does not modify any MLLM parameters and therefore fully preserves the model’s original capabilities.

In addition, we specify the exact baseline model versions used in our experiments. For VLPart~\cite{vlpart}, we use the SwinBase Cascade Mask R-CNN architecture pretrained specifically on the PACO~\cite{ramanathan2023paco} dataset. For X-Decoder~\cite{xdecoder}, we adopt the inference script for OpenVocab Referring Segmentation with the Focal-L backbone. For LISA~\cite{lai2024lisa}, we use the LISA-7B-v1 model, and for GLaMM~\cite{rasheed2024glamm}, we employ the GLaMM FullScope model.
Notably, among these baselines, VLPart~\cite{vlpart} and LISA~\cite{lai2024lisa} directly incorporate PACO data during training. In contrast, X-Decoder~\cite{xdecoder} and GLaMM~\cite{rasheed2024glamm} do not use PACO annotations, but they do train on COCO~\cite{lin2015microsoftcococommonobjects}, which contains many of the same underlying images as PACO.


\section{More Qualitative Results}
\label{supp_qualitative}
\begin{figure*}[tp]
    \centering
    \includegraphics[width=0.95\linewidth]{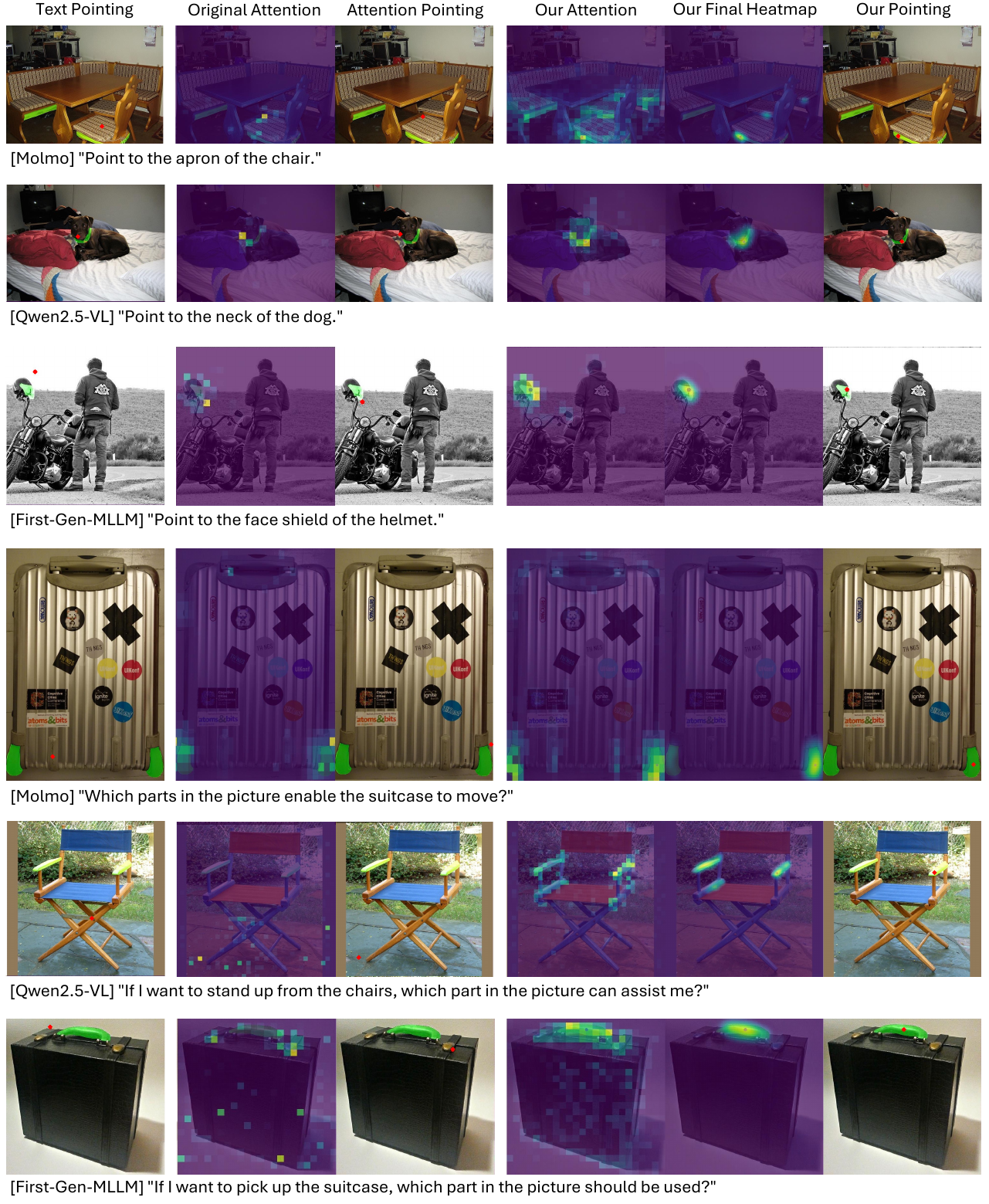}
    \caption{More qualitative results. We compare text pointing, attention pointing, and our proposed method across columns. 
    Each row presents examples from different MLLMs, with the model and text instruction shown below the images. 
    The first three rows correspond to the PACO~\cite{ramanathan2023paco} dataset, while the last three rows correspond to the InstructPart~\cite{wan2024instructpart} dataset.
    In the visualizations, green masks denote ground-truth regions, and red points indicate the predicted point grounding locations.}
    \label{fig:supp_qualitative_ip}
\end{figure*}
We provide additional qualitative results of the PACO~\cite{ramanathan2023paco} and InstructPart~\cite{wan2024instructpart} datasets in Figure~\ref{fig:supp_qualitative_ip}. Notably, some examples contain multiple target instances within the same image, and our attention maps successfully highlight all corresponding regions. Although our current framework performs single-point prediction by selecting the most prominent attention peak, these qualitative observations indicate that extending the method to multi-point outputs is feasible, as further discussed in Sec.~\ref{supp_limitation}.

\section{More Ablation Studies}
\label{supp_ablation}
We follow the setting described in the main manuscript where all the ablation studies are conducted with the First-Gen-MLLM (LLaVA-1.5-7B) on the PACO dataset.

\paragraph{Variants of the Query Synthesis Module.}
\begin{table}
  \caption{Variants of the Query Synthesis Module.} 
  \label{tab:supp_qs_ablation}
  \centering
    \begin{tabular}{lcccc}
    \toprule
    & Avg. Pool & Max Pool & EOS & Ours \\
    \midrule
    Acc. & 0.305 & 0.266 & 0.290 & \textbf{0.463} \\
    \bottomrule
    \end{tabular}
\end{table}
We compare several strategies for utilizing the MLLM’s text features in Table~\ref{tab:supp_qs_ablation}. Instead of using our Query Synthesis (Q-Synth) Module to extract a single query from the full set of text features, one can alternatively apply average pooling or max pooling, or use the feature of the \texttt{[EOS]} token as the text representation. The results show that our proposed Q-Synth Module significantly outperforms these alternatives, demonstrating the effectiveness of our design.

\paragraph{Variants of the SDF-based Supervision.}
\begin{table}
  \caption{Variants of the SDF-based Supervision.}
  \label{tab:supp_sdf_ablation}
  \centering
    \begin{tabular}{lcccc}
    \toprule
    & L2 & Gaussian & Symmetric SDF & Ours \\
    \midrule
    Acc. & 0.433 & 0.436 & 0.454 & \textbf{0.463} \\
    \bottomrule
    \end{tabular}
\end{table}
We compare different loss designs for training the Attention-to-Point (A2P) Decoder in Table~\ref{tab:supp_sdf_ablation}. Instead of supervising the A2P Decoder with the proposed SDF-based Penalty Field, we consider three alternatives: (1) directly using the original binary mask with an L2 loss; (2) using a Gaussian heatmap centered at the innermost point of the mask as supervision; and (3) transforming the mask with the original SDF, without our asymmetric design. The results show that our SDF-based Penalty Field encourages the decoder to produce the most accurate predictions via the point-centric design.

\section{Latency and Model Size}
\label{supp_lantency}
\begin{table}
  \caption{Latency analysis. Our method actually has lower latency than autoregressive text generation.}
  \label{tab:supp_time}
  \centering
    \begin{tabular}{lcc}
    \toprule
    & Text pointing & Ours \\
    \midrule
    LLaVA-1.5-7B & 588.22 ms & \textbf{118.59} ms \\
    \midrule
    Qwen2.5-VL-7B & 1351.19 ms & \textbf{407.78} ms \\
    \bottomrule
    \end{tabular}
\end{table}
For the inference time reported in Table~\ref{tab:supp_time}, we evaluate LLaVA-1.5-7B and Qwen2.5-VL-7B on the same 100 samples using a single NVIDIA A6000 GPU. Counterintuitively, our method actually has \textit{lower} latency than text pointing (autoregressive text generation), since it requires only \textit{a single forward pass} through the MLLM to extract the internal QKV features for point prediction.
In terms of model size, Q-Synth and the A2P Decoder contain 2.3M and 1.7M parameters, respectively, representing a negligible overhead relative to the 7B-parameter MLLM.

\section{Limitations and Future Works}
\label{supp_limitation}
In the main manuscript, we emphasize that our work targets robotic applications in which the model provides a single point per execution step as a high-level signal. A natural extension is enabling the model to output multiple points corresponding to multiple targets within a scene—a capability not currently supported. However, our framework shows promising potential in this direction: the attention patterns produced by the Q-Synth Modules often highlight multiple relevant regions (as shown in Figure~\ref{fig:supp_qualitative_ip}), and the SDF-based penalty field readily generalizes to multiple masks (Figure~\ref{fig:supp_sdf_example}). Incorporating heuristic or learned mechanisms for detecting and separating attention peaks could enable multi-point prediction in future work.

Another limitation is that our method does not explicitly handle scenarios in which no target is present in the image. This issue could be addressed by applying a confidence-based threshold to reject samples with uniformly low attention responses.

Finally, our enhanced attention patterns may also benefit Visual Question Answering (VQA) pipelines, as recent studies~\cite{zhang2025mllms, wu2025flmm} indicate that cropping the image according to MLLM attention improves reasoning via visual chain-of-thought. Exploring how our approach can further strengthen such VQA systems presents an interesting direction for future research.

\end{document}